\documentclass{article} 
\usepackage{iclr2022_conference,times}
\usepackage{graphicx}
\usepackage{xspace,mfirstuc,tabulary}
\usepackage{times}
\usepackage{latexsym}
\usepackage{booktabs}
\usepackage{comment}
\usepackage{xspace}
\usepackage{color, colortbl}
\usepackage{comment}
\usepackage{subcaption}

\newcommand*{\yoruba}{Yor\`ub\'a\xspace}

\usepackage{amsmath,amsfonts,bm}

\def\eqref#1{equation~\ref{#1}}

\def\1{\bm{1}}

\DeclareMathAlphabet{\mathsfit}{\encodingdefault}{\sfdefault}{m}{sl}
\SetMathAlphabet{\mathsfit}{bold}{\encodingdefault}{\sfdefault}{bx}{n}

\usepackage{hyperref}
\usepackage{url}

\title{YOSM: a new \yoruba Sentiment corpus for Movie reviews}

\author{Iyanuoluwa Shode\textsuperscript{1,3}, David Ifeoluwa Adelani\textsuperscript{2,3}, and Anna Feldman\textsuperscript{1} \\
\textsuperscript{1} Montclair State University \\
\textsuperscript{2} Spoken Language Systems (LSV), Saarland University,
Saarland Informatics Campus, Germany\\
\textsuperscript{3} Masakhane NLP \\
\texttt{\{shodei1,feldmana\}@montclair.edu} \\
\texttt{didelani@lsv.uni-saarland.de}
}

\iclrfinalcopy 
\begin{document}

\maketitle

\textbf{Sentiment Analysis} is a popular text classification task in natural language processing. 
It involves developing algorithms or machine learning models to determine the sentiment or opinion expressed in a piece of text. The results of this task can be used by business owners and product developers to understand their consumers' perceptions of their products. Asides from customer feedback and product/service analysis, this task can be useful for social media monitoring~\citep{Martin2021SentimentCI}. One of the popular applications of sentiment analysis is for classifying and detecting the positive and negative sentiments on movie reviews. Movie reviews enable movie producers to monitor the performances of their movies~\citep{IJARCS6536} and enhance the decision of movie viewers to know whether a movie is good enough and worth investing time to watch~\citep{devi_sentiment}. However, the task has been under-explored for African languages compared to their western counterparts, "\textit{high resource languages}", that are privileged to have received enormous attention due to the large amount of available textual data. African languages fall under the category of the low resource languages which are on the disadvantaged end because of the limited availability of data that gives them a poor representation~\citep{Nasim2020SentimentAO}. Recently, sentiment analysis has received attention on African languages in the Twitter domain for Nigerian~\citep{Muhammad2022NaijaSentiAN} and Amharic~\citep{yimam-etal-2020-exploring} languages. However, there is no available corpus in the movie domain. We decided to tackle the problem of unavailability of \yoruba data for movie sentiment analysis by creating the first \yoruba sentiment corpus for Nollywood movie reviews. Also, we  develop sentiment classification models using state-of-the-art pre-trained language models like mBERT~\citep{devlin-etal-2019-bert} and AfriBERTa~\citep{ogueji-etal-2021-small}.  

\textbf{\yoruba Language} is the third most spoken indigenous African language (Eberhard et al., 2020) with over 50 million speakers. Speakers of the \yoruba language can be found in the South-Western region of Nigeria and across the globe. \yoruba is a tonal language that comprises 25 letters.   
Despite its large number of speakers, \yoruba falls under the category of the low resource languages and few NLP datasets that have been developed for the language~\citep{adelani-etal-2021-masakhaner}. Furthermore, there is no record of sentiment analysis research done on Nigerian movies (i.e. \textit{Nollywood}) or even \yoruba movie reviews.

\textbf{Nollywood} is the home for Nigerian movies that depict the Nigerian people and reflect the diversities across Nigerian cultures. A Masterclass staff, Foster in 2022~\footnote{\url{https://www.masterclass.com/articles/nollywood-new-nigerian-cinema-explained}}, claims that four to five movies are released daily by Nigerian movie producers for an estimated audience of fifteen million Nigerians and five million in other African countries. As a result, Nollywood is the second-largest movie and film industry in the world. Despite its capacity, Nollywood movie reviews are scarce. 

\textbf{Data}: Unlike Hollywood movies that are heavily reviewed with hundreds of thousands of reviews all over the internet, there are fewer reviews about Nigerian movies. Furthermore, there is no online platform dedicated to movie reviews originally written in  \yoruba. Most of the reviews are written in English. We collected 1,500 reviews with a balanced set of positive and negative reviews. These reviews were accompanied with ratings and were sourced from three popular online movie review platforms~\footnote{\url{www.imdb.com} , \url{www.rottentomatoes.com},  and \url{https://letterboxd.com/} } - IMDB, Rotten Tomatoes and, Letterboxd. We also collected reviews and ratings from two Nigerian indigenous movie reviews websites~\footnote{\url{www.cinemapointer.com}, and \url{https://nollyrated.com/}} - Cinemapointer and Nollyrated. 
Our annotation focused on the classification of the reviews based on the ratings that the movie reviewer gave the movie. We used a rating scale to classify the positive or negative reviews and defined ratings between 0-4 under the negative (NEG) category while 7-10 were positive (POS). After collecting the data, native speakers of \yoruba that work as professional translators were recruited to manually translate the movie reviews from English to \yoruba. Thus, we have a parallel review dataset in English and Yoruba, and their corresponding ratings. 

As an alternative in the absence of human translation for training, we automatically translate the English reviews to \yoruba using Google Translate machine translation tool, this can be useful for scenarios where there is an absence of training data in \yoruba language. To evaluate the quality of the automatic translation, we compute BLEU score~\cite{papineni-etal-2002-bleu} between human translated sentences and output of Google Translate. We obtained $3.36$ BLEU, which shows the performance of the English-\yoruba MT model is very poor, similar to the observation of \citet{adelani-etal-2021-effect} on Google Translate across several domains. However, we want to evaluate to which extent automatic translations can help when there is an absence of human translations.  
\autoref{tab:data_stat} shows the information about the data sources of the curated \yoruba movie reviews, which we named \texttt{YOSM}. We split \texttt{YOSM} into 800 reviews as training set, 200 reviews as development set and 500 reviews as test set. 

\begin{table}[t]
 \begin{center}
 \resizebox{\columnwidth}{!}{%
 \footnotesize
  \begin{tabular}{l|r|r|rrrrr}
    \toprule
	& \textbf{No. } & \textbf{Ave. Length} & \multicolumn{5}{c}{\textbf{Data source}} \\
    \textbf{Sentiment} & \textbf{Reviews} & \textbf{(No. words)} & \textbf{IMDB} & \textbf{Rotten Tomatoes} & \textbf{LetterBoxd} & \textbf{Cinemapoint} & \textbf{Nollyrated} \\
    \midrule
    positive & 750 & 73 & 402 & 105 & 81 & 101 & 61   \\
    negative & 750 & 63 & 278 & 133 & 101 & 193 & 46   \\

    \bottomrule
  \end{tabular}
  }
  \vspace{-3mm}
  \caption{Data source, number of movie reviews per source, and average length of reviews }
  \label{tab:data_stat}
  \end{center}
\end{table}

\begin{table}[t]
    \begin{subtable}{.3\linewidth}
      \centering
        \begin{tabular}{p{20mm}r}
        \toprule
         & \\
        \textbf{Model} & \textbf{F1-score} \\
        \midrule
        mBERT & $83.2_{\pm1.8}$\\
        mBERT+LAFT & $86.2_{\pm1.3}$ \\
        AfriBERTa & $\mathbf{87.2_{\pm0.6}}$  \\
        
        \bottomrule
        \end{tabular}
        \label{tab:baseline}
        \caption{Benchmark results}
    \end{subtable}%
    \begin{subtable}{.7\linewidth}
      \centering
        
        \begin{tabular}{p{20mm}rrrr}
        \toprule
         & \multicolumn{4}{c}{\textbf{Transfer learning setting}} \\
        \textbf{Model} & \textbf{imdb (en)} & \textbf{en} & \textbf{yo:MT} &  \textbf{en+yo:MT} \\
        \midrule
        mBERT & 61.4 & 61.9 & 71.5 & 74.0 \\
        mBERT+LAFT & \textbf{69.4} & \textbf{71.8} & 76.5 & \textbf{77.9} \\
        AfriBERTa & 64.3 & 69.8 & \textbf{77.1} & \textbf{77.9}\\
        
        \bottomrule
        \end{tabular}
        \label{tab:transfer}
        \caption{Transfer learning (F1-score)}
    \end{subtable} 
    \caption{Benchmark and transfer learning results (F1-score). All results are average over 5 runs except transfer from ``\texttt{imdb}''. }
    \label{tab:results}
\end{table}

\paragraph{Baseline Models} We \textit{fine-tune} two pre-trained language models (PLMs) that have been pre-trained on \yoruba language: mBERT~\citep{devlin-etal-2019-bert} and AfriBERTa~\citep{ogueji-etal-2021-small}. AfriBERTa has been exclusively pre-trained on 11 African languages while mBERT was pre-trained on 104 languages. As an additional baseline model, we make use of a PLM that has been adapted to \yoruba language using language adaptive fine-tuning (LAFT) -- an approach to fine-tune PLM on monolingual texts on a new language using the same masked language model objective as BERT. It has been shown to improve performance on named entity recognition task on \yoruba~\citep{alabi-etal-2020-massive,adelani-etal-2021-masakhaner} and better zero-shot cross-lingual transfer~\citep{pfeiffer-etal-2020-mad}.

\paragraph{Transfer Learning Setting} We examine four transfer learning experiments, (1) \textbf{imdb (en)}: cross-lingual transfer from a large Hollywood movie review dataset (i.e \texttt{IMDB}) with 25,000 samples and zero-shot evaluation on \texttt{YOSM} test set. (2) \textbf{en}: cross-lingual transfer from the English Nollywood movie review -- the size is limited to the 800 samples in the untranslated reviews in our dataset.  (3) \textbf{yo:MT}: trained on machine translation of 800 English Nollywood reviews to \yoruba language. (4) \textbf{en+yo:MT} combined data from the English Nollywood reviews and machine translated reviews. 

\paragraph{Results} \autoref{tab:results} shows the \textbf{baseline results} on PLMs, we obtained very impressive results ($>83$ F1) by training on our small training set (i.e 800 reviews). AfriBERTa and mBERT+LAFT gave better results (more than $86$ F1)  compared to mBERT ($83.2$) since they have been trained exclusively on African languages or adapted using LAFT. For the \textbf{transfer learning results}, we obtained a very good cross-lingual transfer of over ($61$ F1) on all settings. We find the transfer of \textbf{en} to perform better than \textbf{imdb(en)}, an improvement on of $2.4-5.5$ F1 using mBERT+LAFT or AfriBERTa since \textbf{en} captures better the Nollywood domain than \textbf{imdb(en)} that is based on Hollywood reviews. The best transfer approach in the absence of humanly written \yoruba reviews is to train on machine translated reviews (\textbf{yo:MT}) and/or combine with English Nollywood reviews (\textbf{en+yo:MT}), with performance reaching $77.9$ F1. Although, there is a small benefit of combining English and automatically translated \yoruba Nollywood reviews ($0.8-2.5$ F1) to further improve performance over (\textbf{yo:MT}).

\paragraph{Conclusion}
In this paper, we presented the first \yoruba sentiment corpus for Nollywood movie reviews - YOSM that was manually translated from English Nollywood reviews. We perform experiments on this dataset by using the state-of-the-art pre-trained language models and transfer learning approaches which gave us impressive results. The YOSM dataset is publicly available on Github\footnote{\url{https://github.com/IyanuSh/YOSM}}.

\section*{Acknowledgments}
This material is partially based upon work supported by the National Science Foundation under Grant No. 1704113. Also, we thank Cinemapointer for giving us access to use their movie reviews. David Adelani acknowledges the EU-funded Horizon 2020 projects:  ROXANNE under grant number 833635 and COMPRISE (\texttt{http://www.compriseh2020.eu/}) under grant agreement No. 3081705. We appreciate the collective efforts of the following people: Ifeoluwa Shode, Mola Oyindamola, Godwin-Enwere Jefus, Emmanuel Adeyemi, Adeyemi Folusho and Bolutife Kusimo for their assistance during data collection.


\bibliography{iclr2022_conference}
\bibliographystyle{iclr2022_conference}


\end{document}